# Design of a novel convex hull based feature set for recognition of isolated handwritten *Roman* numerals

N. Das[*], S. Pramanik[*], S. Basu[*], P. K. Saha[+], R. Sarkar[*], M. Kundu[*]

*Computer Sc. & Engg. Dept., Jadavpur University,
Kolkata-700032, India.
[+]Electrical and Computer Engineering, The University of Iowa
Iowa City, IA 52242, USA

**Abstract:**
In this paper, convex hull based features are used for recognition of isolated *Roman* numerals using a Multi Layer Perceptron (MLP) based classifier. Experiments of convex hull based features for handwritten character recognition are few in numbers. Convex hull of a pattern and the centroid of the convex hull both are affine invariant attributes. In this work, 25 features are extracted based on different *bays* attributes of the convex hull of the digit patterns. Then these patterns are divided into four sub-images with respect to the centroid of the convex hull boundary. From each such sub-image 25 *bays* features are also calculated. In all 125 convex hull based features are extracted for each numeric digit patterns under the current experiment. The performance of the designed feature set is tested on the standard MNIST data set, consisting of 60000 training and 10000 test images of handwritten Roman using an MLP based classifier a maximum success rate of 97.44% is achieved on the test data.

## 1. Introduction

Handwritten numeral recognition is widely considered as a benchmark problem of Pattern Recognition and Artificial Intelligence. The task of recognition can broadly be classified into two categories, viz., recognition of machine printed text and the recognition of handwritten text. Machine printed characters are uniform in size, position and pitch for any given font. In contrast, handwritten characters are non-uniform, involving variability in the writing style of different individual. Despite these challenges, recognition of handwritten text is a popular research area for many years because of its utility in various commercial applications e.g., automatic postal code recognition, reading amounts from bank cheques, collecting data from filled in forms and so on. In dealing with the problem of recognition of numeric patterns of varying shapes and sizes, selection of a proper feature set is important to achieve high recognition performance. The current research aims to develop a novel convex hull based feature set for effective recognition of isolated handwritten *Roman* numerals.

The convex hull of a point-set is the smallest convex space that contains all the points belongings to the set. For a finite 2D point-set, the convex hull may be defined as the smallest convex polygon containing all the points. In our Present work, we have used *Graham scan algorithm* [6] for computing the convex hull of each numeric pattern. The worst-case complexity of this algorithm for a point-set containing *n* points is $O(n \log n)$. Convex hulls have several useful properties which make them suitable for many recognition and representation tasks.

References related to convex hull based features for handwritten character recognition are few in numbers. C. Gope et al. [1] presented an affine invariant point-set matching technique which measures the similarity between two point-sets by embedding them into an affine invariant feature space, utilizes the convex hull of the point-set to extract affine invariant features. R. Minhas et al. [2] extracted features in images called convex diagonal, convex quadrilateral are used for accurate image registration. Convex diagonals, convex quadrilaterals have attractive properties like easy extraction, geometric invariance and frequent occurrence. P. P. Roy et al. [3] presented a scheme towards recognition of English character in multi-scale and multi-oriented environments. Graphical document such as map consists of text lines which appear in different orientation. The feature set used here is invariant to character orientation. Circular ring and convex hull have been used along with angular information of the contour pixels of the character to make the feature rotation invariant. Circular ring and convex hull have been used to divide a character into several zones and zone wise angular histogram is computed to get higher dimensional feature for better performance. A support vector machine (SVM) classifier has been used for recognition purpose. Yann LeCun et al. [7] described a method uses a linear encoder, and a linear decoder preceded by a sparsifying non-linearity that turns a code vector into a quasi-binary sparse code vector. Given an input, the optimal code minimizes the distance between the output of the decoder and the input patch while being as similar as possible to the encoder output. Learning proceeds in a two-phase EM-like fashion: firstly, to compute the minimum-energy code vector, and secondly to adjust the parameters of the encoder and decoder so as to decrease the energy. The model produces "stroke detectors" when trained on handwritten numerals, and Gabor-like filters when trained on natural image patches. Using the proposed unsupervised method to initialize the first layer of a convolutional network, they achieved an error rate slightly lower than the best reported result on the MNIST dataset. Kai Labusch et al. [10] proposed sparse-coding strategy and a local maximum operation for digit recognition. It first employs the unsupervised sparsenet algorithm to learn a basis for representing patches of handwritten digit images. Then this basis is used to extract local coefficients. In a second step, they apply a local maximum operation in order to implement local shift invariance. Finally, they train a Support-Vector-Machine on the resulting feature vectors to obtain state-of-the-art classification performance in the digit recognition task defined by the MNIST benchmark.

It is true that objects which are very different in shape may have identical convex hulls, and that would be a problem if we were to exclusively use the convex hull attribute as features. In this paper we have extracted different *bays* attributes of the convex hull from the digit images. To extract local information from such images, each such numeric pattern is further divided into four sub-images based on the centroid of its convex hull. After that, new convex hulls are constructed for each such sub-image. By our strategy we have identified the different class of objects though their convex hulls are same. In our method for recognition of handwritten numerals, we have used simple 125 features based on different *bays* attributes of the convex hull using MLP based classifier.

## 2. Computation of convex hull

As discussed earlier, convex hull of any binary pattern may be defined as in this work, we have used Graham scan algorithm for computation of the convex hull of a binary digit pattern. A pseudo code of the said algorithm is discussed below.

**Input:** a set of points **S** = {P = (P.x,P.y)}
    Select the rightmost lowest point $P_0$ in **S.**
    Sort **S** angularly about $P_0$ as a center.
        For ties, discard the closer points.
    Let P[N] be the sorted array of points.
    Push P[0]=$P_0$ and P[1] onto a stack **W**.
    while i < N
    {
        Let $P_{T1}$ = the top point on **W**
        Let $P_{T2}$ = the second top point on **W**
        if (P[i] is strictly left of the line $P_{T2}$ to $P_{T1}$
            Push P[i] onto **W**
            i++   // increment
        else
            Pop the top point $P_{T1}$ off the stack
    }
**Output:** **W** = the convex hull of **S**.

Apart from their computational efficiency, convex hulls are particularly suitable for affine matching as they are affine invariant. In other words, if a point-set undergoes an affine transformation, the convex hull of the point-set undergoes the same affine transformation. Also, convex hulls have local controllability, i.e. they are only locally altered by point insertions/deletions/perturbations. This is a useful property as far as noise tolerance and partial occlusions are concerned. From the green theorem [13] it can be shown that the area A of Convex hull is given by

$$A = \frac{1}{2} \sum_{i=1}^{L}(x_i y_{i+1} - x_{i+1} y_i).$$

L = Number of order vertices, ($x_i$, $y_i$) coordinates of the order vertices forming polygon.
Also, the centroid ( $C_x$, $C_y$ ) of the convex hull can be expressed as

$$c_x = \frac{1}{6A} \sum_{i=1}^{L}(x_i + x_{i+1})(x_i y_{i+1} - x_{i+1} y_i)$$

$$c_y = \frac{1}{6A} \sum_{i=1}^{L}(y_i + y_{i+1})(x_i y_{i+1} - x_{i+1} y_i)$$

it is proved that the centroid ( $C_x$, $C_y$ ) is affine invariant, i.e. the centroid of the affine transformed convex hull is the affine transformed centroid of the original convex hull. Convex hulls of sample digit patterns of *Roman* script, computed by the Graham scan algorithm, are shown in Fig .1.

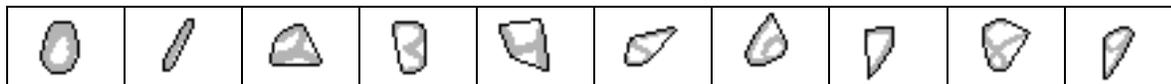

Fig.1.The sample digit images and the corresponding convex hulls are represented in grey and black colors respectively.

The set of pixels inside the convex hull of any object pattern which does not belong to the said object is called the *deficit of convexity*. There may be two types of convex deficiencies viz., region(s) totally enclosed by the object, called *lakes* and region(s) lying between the convex hull perimeter and the object, called *bays*. This is illustrated in Fig.2.

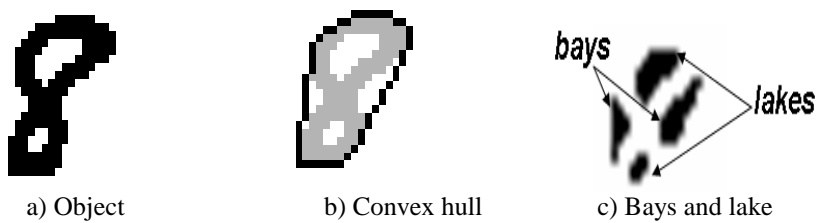

a) Object        b) Convex hull        c) Bays and lake

Fig.2. Illustration of different convex shape descriptors for a sample digit image.

## 3. Design of the feature set

Any object with a non-regular shape may be represented by a collection of its topological components or features. In the current work, we have extracted several such topological features from the convex hull of handwritten *Roman* numerals.

In the current work, 25 features are designed on the basis of different *bays* attributes of the convex hull of handwritten *Roman* numerals. From the top, bottom, right and left boundaries of any image, as shown in Fig.3, column and row wise distances of data pixel from convex hull boundary are calculated as $d_{cp}$. Then the maximum $d_{cp}$, i.e. total no. of rows having $d_{cp} > 0$, Average $d_{cp}$, mean row co-ordinate($r_x$) having $d_{cp} > 0$, total no. of rows having $d_{cp} = 0$, number of visible bays in this direction are computed as six topological features. From the top, bottom, right and left boundaries of the image (6x4=24) such features are calculated. Finally, along the perimeter of the convex hull one more feature is calculated as the total number of perimeter pixels having $d_{cp} = 0$. Fig 4.(a-b) shows the feature extraction techniques for two *Roman* numerals from the Left and Top of the image frames respectively.

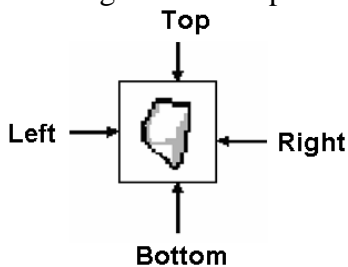

Fig.3. Convex hull of *Roman* digit FOUR. Extract feature from four sides.

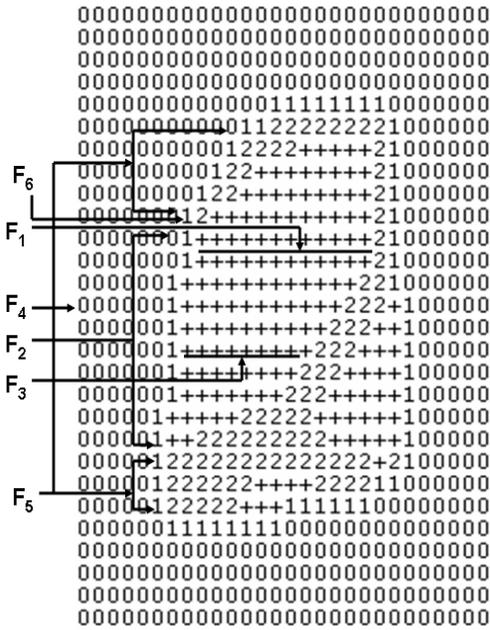

**Row wise from left to right**

| Feature No | Description | Values |
|---|---|---|
| F1 | Maximum dcp | 12 |
| F2 | Total no. of rows having dcp > 0 | 12 |
| F3 | Average dcp | 8.8 |
| F4 | Mean row co-ordinate (rx) having dcp > 0 | 14 |
| F5 | Total no. of rows having (dcp = 0) | 8 |
| F6 | Number of visible bays in this direction | 1 |

Convex hull perimeter feature calculated as, total no. of convex hull pixels having $d_{cp} = 0$ from four sides = 41

( a )

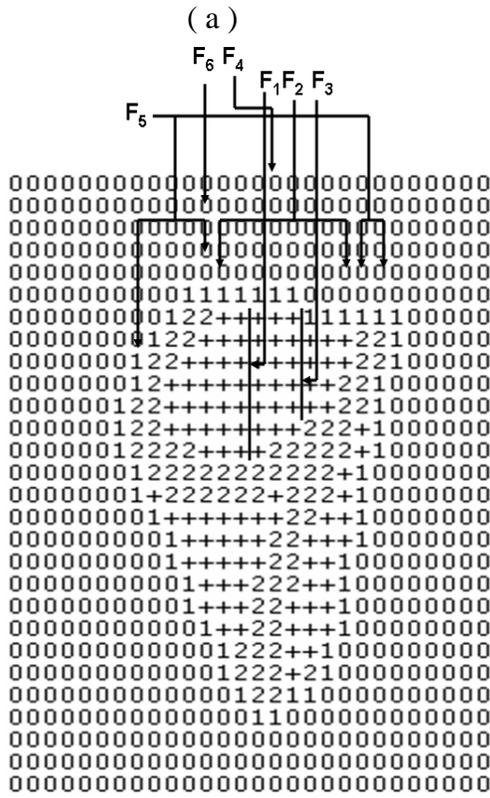

**Column wise from top to bottom**

| Feature No | Description | Values |
|---|---|---|
| F1 | Maximum dcp | 7 |
| F2 | Total no. of rows having dcp > 0 | 8 |
| F3 | Average dcp | 5.4 |
| F4 | Mean column co-ordinate (cx) having dcp > 0 | 15 |
| F5 | Total no. of rows having (dcp = 0) | 7 |
| F6 | number of visible bays in this direction | 1 |

Convex hull perimeter feature calculated as, total no. of convex hull pixels having $d_{cp} = 0$ from four sides = 30

(b)

Fig 4. (a). Binary image of *Roman* Numeral TWO
(b). Binary image of *Roman* Numeral FOUR

The label '1' represents convex hull boundary, label '2' represents digit-pixels, label '+' represents lake and bay regions within the convex hull and label '0' represents background.

As mentioned above, 25 features are extracted from the overall image based on different *bays* attributes of the convex hull. To extract local information, from the digit images, each such numeric pattern is further divided into four sub-images based on the centroid of its convex hull.The convex hulls are then constructed for the digit pixels within each such sub-image for computation of different topological features, as described earlier. 100 such features are computed from the 4 sub-images of each digit pattern. This make the total feature count as 125, i.e. 25 features for the overall image and 100 features in all for the four sub-images.

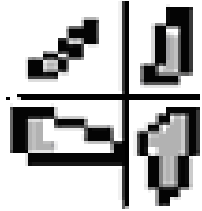

Fig.5. Convex hulls for the sub-images of each quadrant, computed based on the centroid, are shown

## 4. Experimental Result:

For classification of the *Roman* digit patterns of the MNIST dataset into 10 pattern classes, the 125 element feature set, as discussed earlier, is used. MNIST dataset consists of 60000 training and 10000 test images of hand written digit of size 28 x 28 pixels. All these samples are converted to 28 x 28 binary images through thresholding

For the present work, Multi Layer Perceptron(MLP) with one hidden layer is chosen. This is mainly to keep the computational requirement of the same, low without affecting its function approximation capability [14]. To design the MLP for classification of handwritten *Roman* numerals , *Back Propagation* (BP) learning algorithm with learning rate ($\eta$) = 0.8 and momentum term ($\alpha$) = 0.7 is used here for training the classifier with varying number of neurons in its hidden layer. Recognition performances of the MLP, as observed from this experiment are given in Table1. Curves showing the same are also plotted in Fig.6. The recognition rate of the classifier, as recorded on the test data set, initially rises as the number of neurons in its hidden layer is increased and falls after it crosses some limiting value. It reflects the fact that for some fixed training and test sets, generalization ability of an MLP improves as the number of neurons in its hidden layer is increased up to certain limiting value and any further increase in the same thereafter degrades this ability. It is also true for the learning ability recorded on the training set. The phenomenon is called the over-fitting problem. As observed from Table 1 and Fig.6, the best recognition rate on test set is observed as 97.44% with 110 hidden neurons.

| Table 1 | | Neurons | |
|---|---|---|---|
| **No. of Hidden** | **% Success Rate** | 80 | 97.16 |

| | |
|---|---|
| 85 | 97.20 |
| 90 | 97.26 |
| 95 | 97.20 |
| 100 | 97.33 |
| 105 | 97.30 |
| 110 | **97.44** |
| 115 | 97.29 |
| 120 | 97.33 |

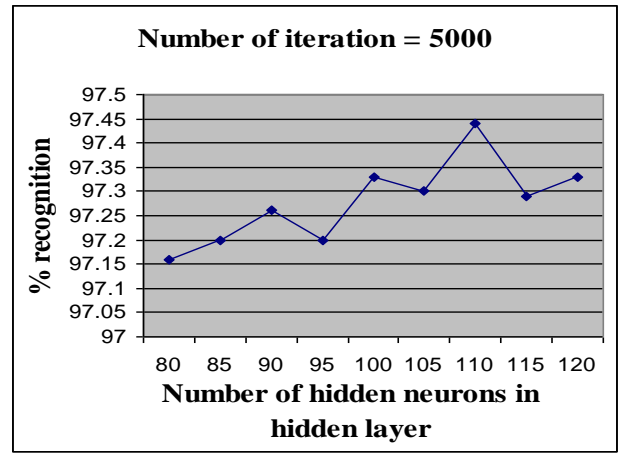

Fig.6.

## 5. Conclusion:

In the current work, an effective convex hull based feature set is designed for recognition of handwritten *Roman* numerals. As observed from the experiments, the current technique achieves a maximum recognition rate of 97.44% on the standard MNIST data set with no rejection. The result is comparable with the ones available in the literatures. The reliability of the system may further be enhanced by introducing rejection criteria and also by further improving the feature set. The designed feature set is novel in the sense that the topological features, extracted from the convex hull boundary of any digit pattern are found to be robust in comparison to the conventional feature sets used in the literature. The affine invariant attributes of this feature set makes it relevant for possible feature combination schemes for enhancement of the overall recognition performance of pattern class. This may be considered as the future direction of the current work.

### Acknowledgements:

Authors are thankful to the "Center for Microprocessor Application for Training Education and Research", "Project on Storage Retrieval and Understanding of Video for Multimedia" of Computer Science & Engineering Department, Jadavpur University, for providing infrastructure facilities during progress of the work.

### References:

**[1].** C. Gope et al., "Affine invariant comparison of point-sets using convex hulls and hausdorff distances", Pattern Recognition, Vol. 40, no. 1, Jan. 2007, pp. 309-320.
**[2].** Rashid Minhas et al., "Invariant Feature Set in Convex Hull for Fast Image Registration", Systems, Man and Cybernetics, Vol. Issue, 7-10 Oct. 2007 pp.1557 - 1561
**[3].** P.P.Roy et al., "Convex Hull based Approach for Multi-Oriented Character Recognition from Graphical Documents", 2008
**[4].** S.Basu et al., "Handwritten 'Bangla' Alphabe recognition using an MLP based classifier", NCCPB-2005, Bangladesh, pp.285-291.

**[5].** Z.Yang et al., "Image Registration and Object Recognition Using Affine Invariants and Convex Hulls", IEEE Transactions on Image Processing, vol. 8, no. 7, july 1999

**[6]** Laura Vyšniauskaitė et al., "A Priori Filtration Of Points For Finding Convex Hull", Tede. 2006, Vol XII, No 4, 341–346.

**[7]** M. Ranzato, C. Poultney, S. Chopra, and Y. LeCun, "Efficient learning of sparse representations with an energy-based model," in Advances in Neural Information Processing Systems 19, B. Schölkopf, J. Platt, and T. Hoffman, Eds. Cambridge, MA: MIT Press, 2007, pp. 1137–1144.

**[8]** P. Y. Simard, D. Steinkraus, and J. C. Platt, "Best Practices for Convolutional Neural Networks Applied to Visual Document Analysis,"in ICDAR '03: Proceedings of the Seventh International Conference on Document Analysis and Recognition. Washington, DC, USA: IEEE Computer Society, 2003, p. 958.

**[9]** F. Lauer, C. Y. Suen, and G. Bloch, "A trainable feature extractor for handwritten digit recognition," Pattern Recogn., vol. 40, no. 6, pp. 1816–1824, 2007.

**[10]** Kai Labusch et al.,"Simple Method for High-Performance Digit Recognition Based on Sparse Coding".

**[11]** D. Decoste and B. Schölkopf, "Training Invariant Support Vector Machines," Mach. Learn., vol. 46, no. 1-3, 2002, pp. 161–190.

**[12]** F. Lauer, C. Y. Suen, and G. Bloch, "A trainable feature extractor for handwritten digit recognition," Pattern Recogn., vol. 40, no. 6, pp. 1816–1824, 2007.

**[13]** E. L. Lady , http://www.math.hawaii.edu/~lee/calculus/green.pdf, February 14, 2000

**[14]** N. J. Nilson, "Principles of Artificial Intelligence," Springer-Verleg, pp. 21-22.

## Authors' Biography

**Nibaran Das** received his B.Tech degree in Computer Science and Technology from Kalyani Govt. Engineering College under Kalyani University, in 2003. He received his M.C.S.E degree from Jadavpur University, in 2005. He joined J.U. as a lecturer in 2006. His areas of current research interest are OCR of handwritten text, Bengali fonts, biometrics and image processing. He has been an editor of Bengali monthly magazine "Computer Jagat" since 2005.

**Sandip Pramanik** received the B.Tech degree in computer science & engineering in 2006 from Kalyani Govt. Engg. Collage. He worked as Research Associate at Indian Institude of Technology Kanpur for one year. He is pursuing Master of Computer Science & Engineering from Jadavpur University, Kolkata. His area of current research interest are pattern classification and character recognition

**Subhadip Basu** received his B.E. degree in Computer Science and Engineering from Kuvempu University, Karnataka, India, in 1999. He received his Ph.D. (Engg.) degree thereafter from Jadavpur University (J.U.) in 2006. He joined J.U. as a senior lecturer in 2006. His areas of current research interest are OCR of handwritten text, gesture recognition, real-time image processing.

**Punam Kumar Saha** received his BCSE, MCSE degrees from Jadavpur University, in 1987, 1989 respectively. He received his Ph.D. (Engg.) degree thereafter from Indian Statistical Institute in 2006. He is a Associate Professor of department of Electrical and Computer Engineering, The University of Iowa, His current research interest include Tensor scale-based image analysis; digital topology and geometry; fuzzy distance transform, virtual bone biopsy; fuzzy connectedness-based object segmentation; etc

**Ram Sarkar** received his B.Tech degree in Computer Science and Engineering from University of Calcutta, in 2003. He received his M.C.S.E degree from Jadavpur University, in 2005. He joined J.U. as a lecturer in 2008. His areas of current research interest are document image processing, line extraction and segmentation of handwritten text images.

**Mahantapas Kundu** received his B.E.E, M.E.Tel.E and Ph.D. (Engg.) degrees from Jadavpur University, in 1983, 1985 and 1995, respectively. Prof. Kundu has been a faculty member of J.U since 1988. His areas of current research interest include pattern recognition, image processing, multimedia database, and artificial intelligence.

**Author Address:** Nibaran Das**,** Computer Sc. & Engineering. Department, Jadavpur University, Kolkata-700032, India , **Phone No:** +913325139162, **Email:** nibaran@cse.jdvu.ac.in